\documentclass[preprint,12pt,3p]{elsarticle}

\usepackage{amssymb}
\usepackage{todonotes}
\usepackage[authoryear]{natbib}
\usepackage{hyperref}
\hypersetup{
    colorlinks=true,
    linkcolor=blue,
    filecolor=blue,      
    urlcolor=blue,
    citecolor=blue,
}

\journal{Elsevier}

\begin{document}

\begin{frontmatter}

\title{Simulation and Optimisation of Air Conditioning Systems using Machine Learning}

\cortext[cor1]{Corresponding author. Postal Address: Faculty of Information Technology, P.O. Box 63 Monash University, Victoria
3800, Australia. E-mail address: christoph.bergmeir@monash.edu}

\author{Rakshitha Godahewa}

\author{Chang Deng}

\author{Arnaud Prouzeau}

\author{Christoph Bergmeir \corref{cor1}}

\address{Faculty of Information Technology, Monash University, Melbourne, Australia}

\begin{abstract}
In building management, usually static thermal setpoints are used to maintain the inside temperature of a building at a comfortable level irrespective of its occupancy. This strategy can cause a massive amount of energy wastage and therewith increase energy related expenses. 
This paper explores how to optimise the setpoints used in a particular room during its unoccupied periods using machine learning approaches. We introduce a deep-learning model based on Recurrent Neural Networks (RNN) that can predict the temperatures of a future period directly where a particular room is unoccupied and by using these predicted temperatures, we define the optimal thermal setpoints to be used inside the room during the unoccupied period. 
We show that RNNs are particularly suitable for this learning task as they enable us to learn across many relatively short series, which is necessary to focus on particular operation modes of the air conditioning (AC) system.
We evaluate the prediction accuracy of our RNN model against a set of state-of-the-art models and are able to outperform those by a large margin. We furthermore analyse the usage of our RNN model in optimising the energy consumption of an AC system in a real-world scenario using the temperature data from a university lecture theatre. Based on the simulations, we show that our RNN model can lead to savings around 20\% compared with the traditional temperature controlling model that does not use optimisation techniques.
\end{abstract}

\end{frontmatter}

\section{Introduction}
\label{sec:intro}
A limited number of energy resources, a rapid growth of population and an increment of energy consumption \citep{ref_1} make energy optimisation a pressing need for nowadays societies. Improving energy efficiency can be an important contribution towards solving the problem of energy shortage. Energy efficiency can also bring substantial economic benefits.
In urban areas, more than 40\% of the energy is consumed from buildings \citep{ref_2} where Heating, Ventilation and Air Conditioning (HVAC) systems consume a large portion of energy in many commercial buildings \citep{ref_3}. The purpose of an HVAC system is to maintain proper and comfortable thermal conditions inside a building. The common approach of operating the HVAC system is by using static thermal setpoints where the inside temperature level is always maintained within a predefined temperature limit. However, this strategy is only beneficial during occupied periods of a building. Maintaining the same temperature level during unoccupied periods may be unnecessary and lead to energy wastage. 

\begin{figure}[htb]
\centering
  \includegraphics[width=\textwidth]{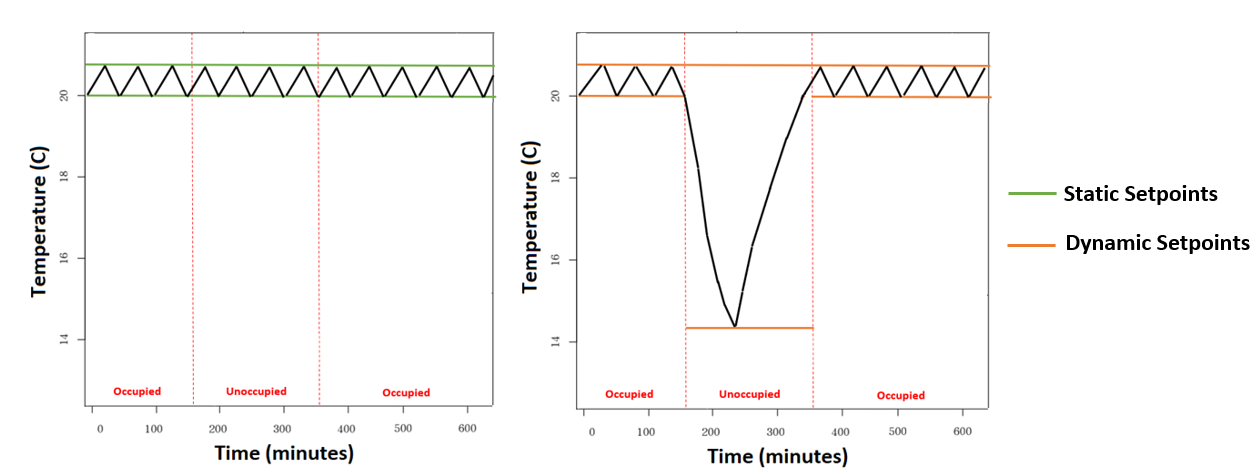}
  \caption{(left) Static thermal setpoints. The system always keeps the inside temperature in between two specific temperature levels irrespective of room occupancy. (right) Dynamic thermal setpoints. The temperature is allowed to drop to a certain level during unoccupied periods. As a result, large amounts of energy can be saved during periods where the room is unoccupied.}~\label{fig:setpoints}
\end{figure}

Dynamic thermal setpoints are a possible approach to control the inside temperature of a building more efficiently than with static thermal setpoints. The approach can define different setpoints for the occupied and unoccupied periods and hence be more energy efficient by allowing the inside temperature to be closer to the outside temperature during unoccupied periods. Figure~\ref{fig:setpoints} illustrates the concept of static and dynamic thermal setpoints. 

For the use of dynamic setpoints and for energy efficiency optimisation, it is highly beneficial to know in advance when rooms are occupied, and it is necessary to model and predict inside temperatures, so that the HVAC system has enough time to bring a room back to a comfortable level at the time it is needed.
%
%
Consequently, there is a body of literature available about forecasting of inside temperatures and energy consumption within buildings. In particular, researchers have used several machine learning technologies in this space, such as \textit{Multiple Linear regression (MLR)}, \textit{Support Vector Machines (SVM)}, \textit{Random Forests (RF)} and \textit{Feed-Forward Neural Networks (FFNN)} \citep{ref_50, ref_32, ref_33, ref_34, ref_39,ref_40, ref_41}.

In the general forecasting literature, Neural Networks (NN) have been traditionally largely disregarded and found not to be competitive. For example, \citet{ref_58} demonstrate that statistical methods are more capable of providing accurate forecasts compared to NNs and other machine learning approaches.
However, this notion is changing recently and there is now a trend towards using machine learning techniques, and in particular \textit{Recurrent Neural Networks (RNN)}, successfully with many forecasting problems \citep{ref_51}. An RNN is a special type of NN that is highly suitable for sequence modelling as it can address the temporal order and temporal dependencies of a sequence \citep{ref_5}. RNNs have also been used to address some real-world building management related forecasting problems \citep{ref_43, ref_44}. RNNs are quite suitable for modelling in our scenario due to their capability of addressing the cold-start problem as explained in Section \ref{sec:methodology}.


The concept of dynamic thermal setpoints has been used to optimise the behaviour of HVAC systems. Recent work in this area uses the predictions of building management related factors such as occupancy and percentage of dissatisfied building occupants to dynamically change the setpoints of buildings \citep{ref_47, ref_48}. 



Although researchers have considered RNNs to solve many real-world forecasting problems related to building management, to the best of our knowledge, there is no prior work on using RNNs for indoor temperature forecasting, especially if these forecasts are then used to optimise the energy consumption of HVAC systems. We examine the merits of recent advances in the field of forecasting, namely the good performance of RNNs and their applicability in such an optimisation system.  
Our paper has the following main contributions.

\begin{figure}[htb]
\centering
  \includegraphics[width=\textwidth]{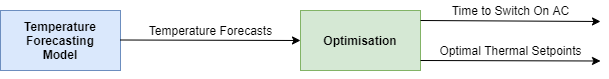}
  \caption{Overall process of setpoint optimisation during unoccupied periods }~\label{fig:overall}
\end{figure}

\begin{enumerate}

\item We propose a deep-learning model based on RNNs that can forecast the future indoor temperatures of a particular building as a function of current inside temperature and outside temperature. We compare the prediction accuracy of our RNN model with a set of baseline machine learning benchmark models, namely RF, SVM, MLR and FFNN. We show that our RNN model is more accurate in forecasting temperatures than the benchmark models.

\item We then use the RNN model to simulate the thermal behaviour inside a particular room and based on that, we propose an optimisation approach to find the optimal time point to switch on the Air Conditioning System (AC) during an unoccupied period of that room such that it uses a minimal amount of energy to heat or cool the room. 
Figure \ref{fig:overall} represents the overall procedure that is used to optimise the setpoints during unoccupied periods. As the setpoints depend on the predicted temperatures, it is crucial for the system to have an accurate forecasting engine at its core.
    
\item We present a quantitative comparison of the relative performance of the optimised RNN model and the second most accurate model, an SVM model, against a traditional AC system in a case study of a university lecture theatre. We show that the optimised models can achieve considerable energy savings compared with the traditional AC system in terms of the amount of energy usage. In particular, relying on the RNN model for predictions leads to an estimated 20\% of energy saving according to our simulations.
\end{enumerate}

The remainder of this paper is organized as follows: Section \ref{sec:relw} reviews the related work. Section \ref{sec:methodology} describes our proposed temperature prediction model based on RNNs and the setpoint optimisation procedure. Section \ref{sec:experiments} summarises our experimental framework and model evaluation results based on prediction accuracy. Section \ref{sec:case_study} presents a case study for using our proposed RNN model to optimise the energy consumption of an HVAC system in a real-world application. Finally, Section \ref{sec:conclusion} concludes the paper.

\section{Related Work}
\label{sec:relw}

In this section we review the existing state-of-the-art temperature and energy prediction models, and HVAC optimisation techniques.
The literature provides many examples of using machine learning technologies to forecast the inside temperatures and energy consumption within buildings. As stated earlier, popular machine learning models in this area are MLR, SVM, RF and FFNN. \citet{ref_25} propose an MLR model to predict the inside thermal behaviour for three separate rooms. They compare the predicted results with the simulation of the COMFIE software \citep{ref_50} where the simulated results are similar to the predictions. \citet{ref_26} further propose a similar MLR model to predict the inside temperatures in summer and winter separately. 

\citet{ref_32} use an approach based on SVM and FFNN to predict the maximum athmospheric temperature of a particular location. \citet{ref_33} use an SVM to predict the daily maximum temperature which is then used to forecast a daily maximum energy consumption. \citet{ref_34} establish a model based on SVM and FFNN to predict the outside temperature in different locations in Australia and New Zealand. Their results show that the SVM has better prediction accuracy compared to the FFNN. Furthermore, \citet{ref_55} proposes a model based on the abductive networks approach to predict hourly temperature forecasts.

Researchers have also used RFs to predict the energy expenditure in buildings. \citet{ref_39} compare FFNNs and RFs in predicting the energy consumption in buildings where both the FFNN and the RF demonstrate similar performance. \citet{ref_40} establish an RF and a gradient boosted tree algorithm to predict the energy consumption in buildings where the gradient boosted tree algorithm achieves better accuracy in prediction. \citet{ref_41} use an RF to predict hourly electricity usage, with good results.

A series of temperatures can be also considered as a sequence/time series. Therefore, it is possible to apply time-series forecasting techniques such as RNN, Exponential Smoothing State Space Model \citep[ETS, ][]{ref_19}, and Autoregressive Integrated Moving Average \citep[ARIMA, ][]{ref_20} to predict a set of future temperatures. 

RNNs are a particularly promising forecasting model.
They recently contributed to the winning solution of the M4 forecasting competition \citep{ref_49}, and RNNs can be trained globally across multiple time series leveraging cross-series information. RNNs are also obtaining competitive results in real-world applications nowadays. For example, \citet{ref_12} use RNNs for sales demand forecasting and are able to outperform statistical benchmarks and a production system. 
 
RNNs have been used to address many building management related forecasting problems in the recent past. \citet{ref_42} use RNNs to predict building energy usage based on hourly data recorded at an engineering center. Further, \citet{ref_43} propose a heat load prediction approach using RNNs where the method provides more accurate forecasts compared to a layered NN. RNNs have also been used for power forecasting \citep{ref_44}, electricity price forecasting \citep{ref_57}, equipment gear life forecasting \citep{ref_60}, dynamic system identification \citep{ref_56} and anomalous behaviour detection of superconducting magnets \citep{ref_59}. Although researchers have considered RNNs to solve many real-world forecasting problems related to building management, to the best of our knowledge, we are the first to use RNNs for indoor temperature forecasting.

Researchers have used the concept of dynamic thermal setpoints to optimise the behaviour of HVAC systems in the past. \citet{ref_48} establish an optimisation model to control the status of the HVAC system dynamically by predicting the room occupancy. Their results show that their system can save up to 21\% of energy on average. \citet{ref_46} evaluate two strategies for changing the temperature setpoints in office buildings: static intervention and dynamic intervention where static intervention raises the setpoints by 1C and dynamic intervention involves load shifting. The results show that dynamic intervention has better performance with a 6.3\% reduction of energy consumption compared to static intervention. Similarly, \citet{ref_47} propose an optimisation model to minimise a proposed cost function that can adjust the temperature setpoints in 15-minute intervals. The cost function contains the Predicted Percentage of Dissatisfied building occupants (PDD), power of the HVAC system, electricity price and $CO{_2}$ index. Their results demonstrate that their proposed optimisation model can save 10\% - 15\% of energy consumption of the HVAC system. The success of previous studies in reducing the energy consumption by using dynamic thermal setpoints encourages us to use such an approach to optimise the HVAC system.

As the literature lacks approaches to use the indoor temperature forecasts to change the setpoints dynamically, we propose a deep-learning temperature forecasting model based on RNNs where the temperature predictions subsequently enable us to dynamically change the setpoints of particular rooms especially during unoccupied periods.

\section{Methodology}
\label{sec:methodology}

The main goal of our research is to optimise the HVAC system by finding the optimal thermal setpoints to be used inside a particular room during its unoccupied periods. For that, first we forecast the indoor temperatures of a future unoccupied period using an RNN model. Then, the forecasts are used to define the optimal thermal setpoints. 
The details of our proposed prediction model and its optimisation procedure are described in the following.




\subsection{Prediction model}
\label{sec:optimisation}

The overall system has four possible states: active heating or cooling, and passive heating or cooling.
In winter, when the outside temperature is lower than the inside temperature, the room will switch between active heating and passive cooling states when the HVAC is switched on and off, respectively. In summer, the system switches accordingly between active cooling and passive heating. 

\begin{figure}[htb]
\centering
  \includegraphics[width=\textwidth]{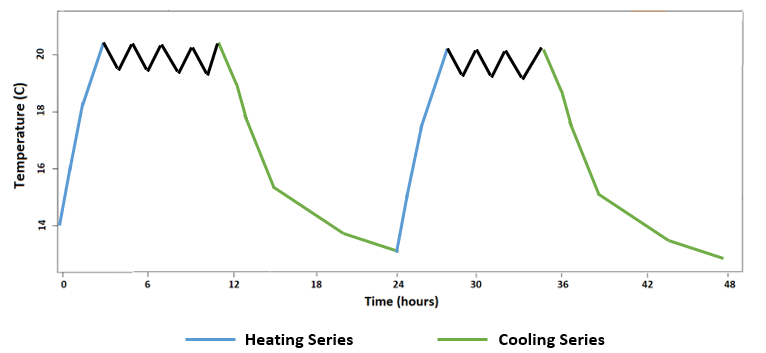}
  \caption{Heating and cooling parts of a temperature series. As these parts show a quite different behaviour, we use different models to fit them. The models need to be able to train on many relatively short heating or cooling series, respectively.}~\label{fig:heating_cooling_series}
\end{figure}

As the underlying mechanisms and the dynamics in the measured temperatures are quite different, we use a separate RNN model to forecast the inside temperatures for each of the four states.
Each such model is trained only with the parts of the time series that correspond to their respective state. E.g., the model for active heating is only trained with periods of active heating. For an illustration, see Figure \ref{fig:heating_cooling_series}. In this way, the model trains effectively with many (relatively short) time series. Therefore, we train our models as global forecasting models that can learn across multiple time series during the training process \citep{ref_53}. 
Additionally to the current inside temperature and current outside temperature, we also provide future outside temperatures obtained from weather forecasts as inputs to the models.

Traditional univariate forecasting models would be of limited use in our setup, as we deal in many situations with a so-called cold-start problem, i.e., predicting series with very short history or no history at all, which corresponds to predicting directly after, e.g., heating has been switched on, in our application.
In particular, our RNN models predict one future inside temperature at a time, and feed the prediction back into the model to predict the next inside temperature. RNNs are especially suitable to model this situation as they memorize the previous outputs using their feedback loops and use them when providing a new forecast. Other machine learning models such as FFNNs and (causal) convolutional networks require a potentially large number of inputs to produce accurate forecasts and hence, they are of limited use in our situation.

We note that our approach can also be considered a clustering approach where two clusters of time series having different characteristics are used to train separate RNN models, as presented by \citet{ref_9}.



\subsubsection{Recurrent Neural Networks with Long Short-Term Memory Cells}
\label{sec:rnn}




The Elman RNN unit \citep[ERNN,][]{ref_5}, the Long Short-Term Memory cell \citep[LSTM,][]{ref_6}, and the Gated Recurrent Unit \citep[GRU,][]{ref_7} are some of the commonly used RNN units for time series forecasting. Out of those, we use the LSTM in our work, which is a popular choice due to its capability of capturing long-term dependencies while addressing vanishing/exploding gradient problems \citep{ref_13}.
%
%
An LSTM cell contains two memory components, a short-term and a long-term memory component, which correspond with its two states, the hidden state and the cell state. 
Furthermore, an LSTM cell contains three gates: input gate, forget gate, and output gate. The input and forget gates determine the amount of past information to be saved in the current cell state and the amount of information to be passed forward to future time points.
In our model, we use an LSTM cell with peephole connections \citep{ref_8} which considers the previous cell state in the updating process of input and forget gates, and is usually more accurate than the vanilla LSTM. 
For further details of LSTMs for forecasting, we refer to \citet{ref_10}.

\subsubsection{RNN Model Architecture}
\label{sec:direct_architecture}

\begin{figure}[htb]
\centering
  \includegraphics[width=\textwidth]{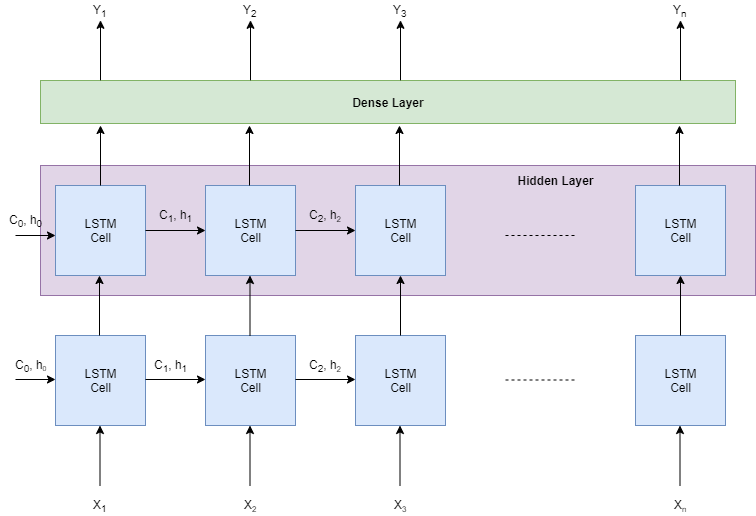}
  \caption{Stacked Architecture used in our RNN Model}~\label{fig:stack}
\end{figure}

We use the stacked architecture for RNN models as suggested by \citet{ref_10} and \citet{ref_9}. Figure \ref{fig:stack} represents the unfolded version of an RNN across time with one hidden layer. In this case, each time step corresponds to an LSTM cell. In Figure \ref{fig:stack}, we denote $X_t$ as the input to the LSTM cell at time step $t$, ${Y}_t$ as the output of the dense layer corresponding to each LSTM cell at time step $t$ and $h_t$ and $C_t$ as the hidden state and cell state of the LSTM cell at time step $t$, respectively. We note that $X_t$ and $Y_t$ are vectors that can contain multiple data points. The feedback loops between LSTM cells support the model in carrying the states of the cells, namely hidden states and cell states, to the future time points.
In the stacked architecture, multiple layers can be placed on top of each other. The inputs are taken at the bottom layer and the corresponding output will be propagated to the next layer. The cell functions calculate the outputs according to the current weights. There can be several hidden layers within the network. 
%
%
To convert the dimension of the cell output to the amount of forecasts needed for the selected forecast horizon, the cell outputs are fed into a dense layer, a neural layer trained together with the RNN. The output of the final LSTM cell $Y_n$ contains the expected forecasts of a particular time-series. 
The cell dimension and number of hidden layers are externally tuned hyperparameters as described in Section \ref{sec:direct_hyperparameters}.

The model training process uses the errors of all time steps until the end of the sequence to calculate an accumulated training error. Equations \ref{eq4} and \ref{eq5} represent the formulas for calculating the error per time step ($e_{t}$), and the final accumulated error ($E$), respectively, where $Z_t$ is the actual output vector at time step $t$. The calculated accumulated error is then used with the Backpropagation Through Time (BPTT) process to update the weights of the RNN cells.

\begin{equation}
    e_{t} = Z_{t} - {Y}_t\label{eq4}
\end{equation}

\begin{equation}
    E = \sum_{t=1}^{T} e_{t}\label{eq5}
\end{equation}

The recent literature suggests to apply a moving window scheme to split the time series into a set of input and output vectors and use them to train the RNN model \citep{ref_12, ref_10}. 
However, as stated before, in our particular application we are frequently in the situation of a cold-start problem after switching states, so that we can effectively use only the current inside and outside temperatures as direct inputs into our model, and not a window of lagged inputs.
Therefore, instead of input and output windows, we predict one future inside temperature at a time, and feed the predictions back into the model in an iterative way until we have predicted the full unoccupied period.

\subsubsection{Hyperparameter Tuning}
\label{sec:direct_hyperparameters}

Each RNN model contains a set of hyperparameters that need to be tuned. For this, we use the last value of each temperature series as a validation set. Then, we use the Sequential Model based Algorithm Configuration (SMAC) optimisation method \citep{ref_14} to automatically identify the optimal values for the hyperparameters given pre-defined ranges for the hyperparameter selection. Table~\ref{tab:hyperparameters} shows the hyperparameters and the initial ranges we use in our experiments.

\begin{table}
\begin{center}
\begin{tabular}{ l|c|c } 
\hline
\textbf{Parameter} & \textbf{Minimum Value} & \textbf{Maximum Value}\\ 
\hline
Cell Dimension & 10 & 15 \\ 
Maximum Number of Epochs & 2 & 25  \\ 
Maximum Epoch Size & 2 & 10 \\ 
Mini-batch Size & 1 & 15  \\ 
Number of Hidden Layers & 1 & 2 \\ 
L2-Regularization Weights & 0.0001 & 0.0008 \\ 
Std. of Random Normal Initializer & 0.0001 & 0.0008 \\ 
Std. of Gaussian Noise & 0.0001 & 0.0008 \\ 
\hline
\end{tabular}
\caption{Initial Parameter Ranges used with Hyperparameter Tuning of RNN Model}
\label{tab:hyperparameters}
\end{center}
\end{table}


\subsubsection{Learning algorithm}

We use the COntinuous COin Betting \citep[COCOB,][]{ref_15} algorithm as the learning algorithm. Unlike other such algorithms including Adam \citep{ref_16} and Adagrad \citep{ref_17}, COCOB does not require the learning rate as a hyperparameter to train the models as it automatically chooses an optimal learning rate. Recent work by \citet{ref_10} proposes to use COCOB as the learning algorithm for RNN training processes in time series forecasting, based on experiments on six benchmark datasets.

\subsection{Optimisation method}


We assume that the passive behaviour of a room is that the inside temperature strives towards the outside temperature in an exponentially decaying way. I.e., the closer the inside temperature is already to the outside temperature, the longer it takes for the inside temperature to change more towards the outside temperature.

\begin{figure}[htb]
\centering
  \includegraphics[width=14cm, height=7cm]{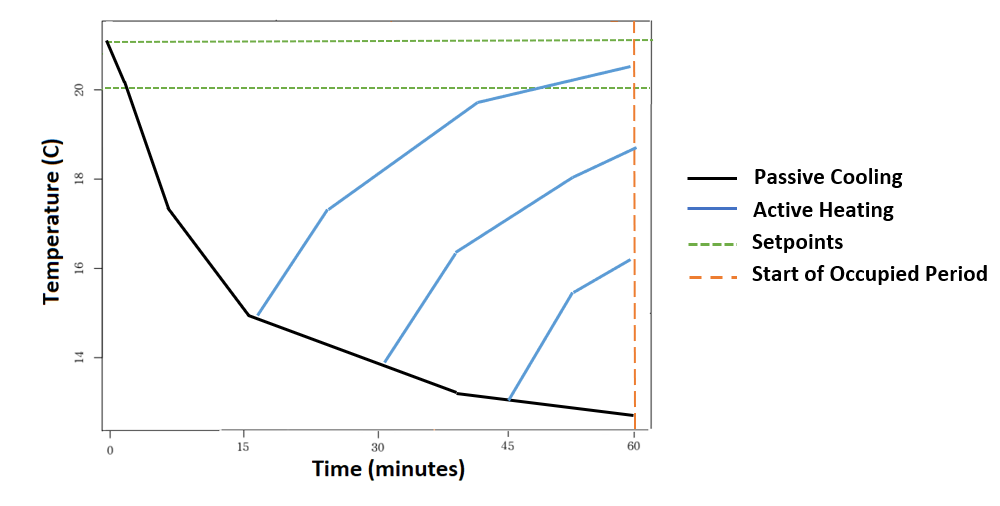}
  \caption{HVAC optimisation procedure for a 60 minutes unoccupied period considering passive cooling and active heating behaviours. According to the predictions, the HVAC system should be switched on 45 minutes before occupying the room to bring the inside temperature to a comfortable level.}~\label{fig:hvac_optimisation}
\end{figure}

\begin{figure}
\centering
  \includegraphics[width=0.95\textwidth]{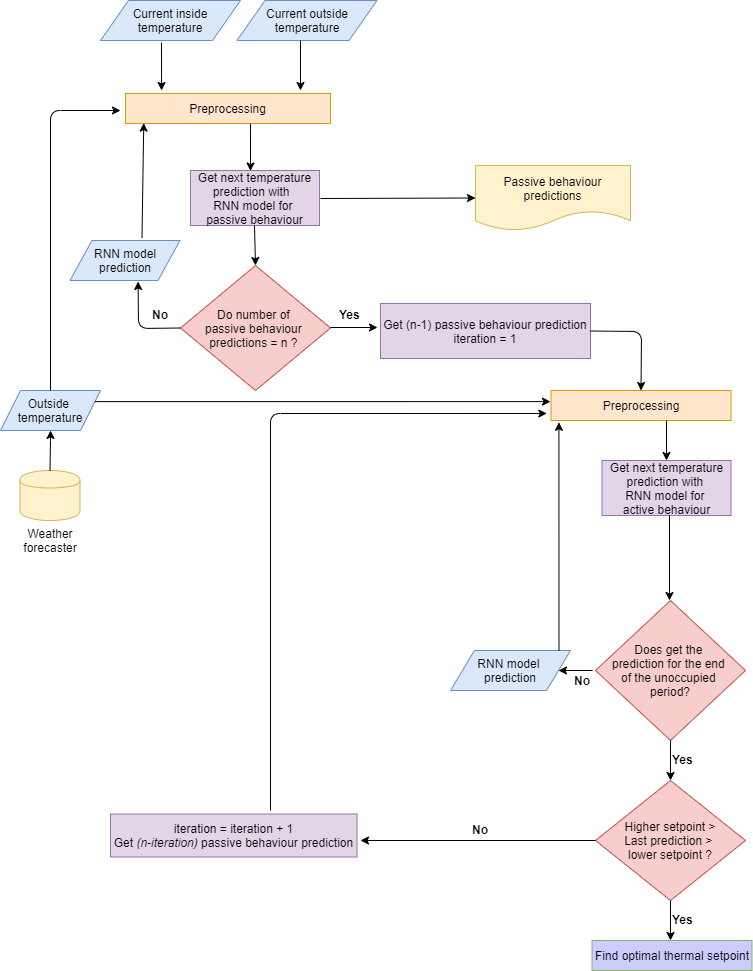}
  \caption{Setpoint Optimisation Procedure Using the Prediction Models. Here, $n$ is the Total Predictions Fit with the Unoccupied Period }~\label{fig:setpoint_optimisation}
\end{figure}

This assumption allows us to implement a straightforward optimisation procedure as follows. The best way to minimise energy consumption of the HVAC system is to leave it in a switched-off state as long as possible until the inside temperature gets closer to the outside temperature. Only when the room is occupied again, there should be a comfortable inside temperature level for the occupants. Therefore, we need to be able to predict the passive change in inside temperature, and the time it will take for the HVAC system to bring back the temperature to a comfortable level before occupation starts. In particular, we proceed as follows.
As the first step, we predict the temperatures for the whole unoccupied period assuming the passive temperature behaviour during this period. Then, we predict the active temperature behaviour of the room, using the respective prediction engine, from different starting points. We start from the last time point of the unoccupied period, and then produce a prediction successively going back one point at a time, until we reach a point from which the predicted temperature at the end of the unoccupied period is within the setpoints, if the HVAC would be switched on at this point in time. The procedure is illustrated in Figure~\ref{fig:hvac_optimisation} considering passive cooling and active heating behaviours.
Figure~\ref{fig:setpoint_optimisation} illustrates the overall setpoint optimisation procedure used with the temperature prediction models.

%
%

\section{Experimental Framework and Results}
\label{sec:experiments}
In this section, we present our experimental setup and results on a real-world temperature dataset. This section describes the dataset, data preprocessing techniques and benchmarks used for model evaluation.

\subsection{Dataset}
\label{sec:dataset}

\begin{figure}[htb]
\centering
  \includegraphics[width=\textwidth]{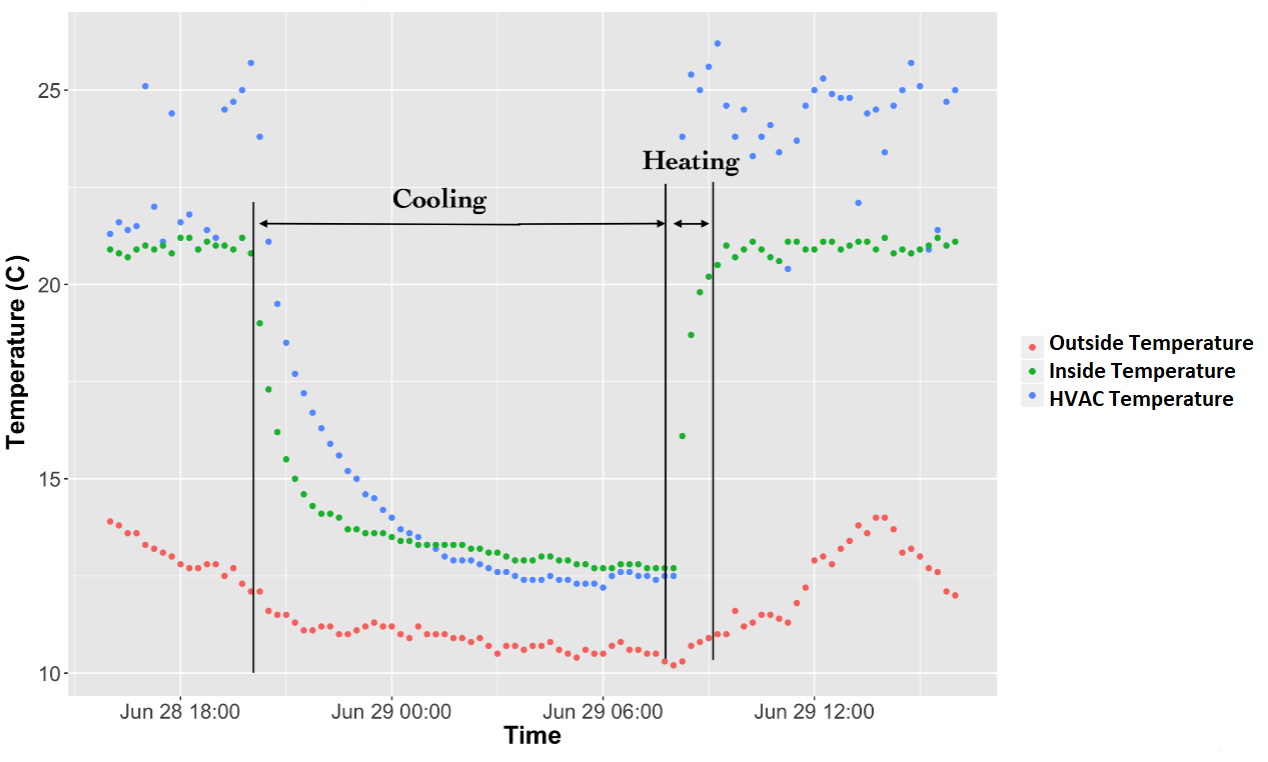}
  \caption{Passive cooling and active heating periods between June 28 and June 29. The lecture theatre cools down during the night when the HVAC is switched off, and then heats up in the morning when the HVAC is switched on.}~\label{fig:TempBehavior}
\end{figure}

We use temperature readings from a university lecture theatre in our experiments. The dataset contains 15-minutely readings between 27/06/2017 and 02/10/2017, 9408 rows in total. Each row contains a timestamp, inside temperature, outside temperature (both measured in degree Celsius), status of the AC (on/off), and a setpoint.
As an example, Figure~\ref{fig:TempBehavior} shows the temperature behaviour on two selected winter days (Australian winter), along with the active (heating) and passive (cooling) periods between these days.

\subsection{Data Preprocessing}
\label{sec:preprocessing}



\begin{figure}[htb]
\centering
  \includegraphics[width=\textwidth]{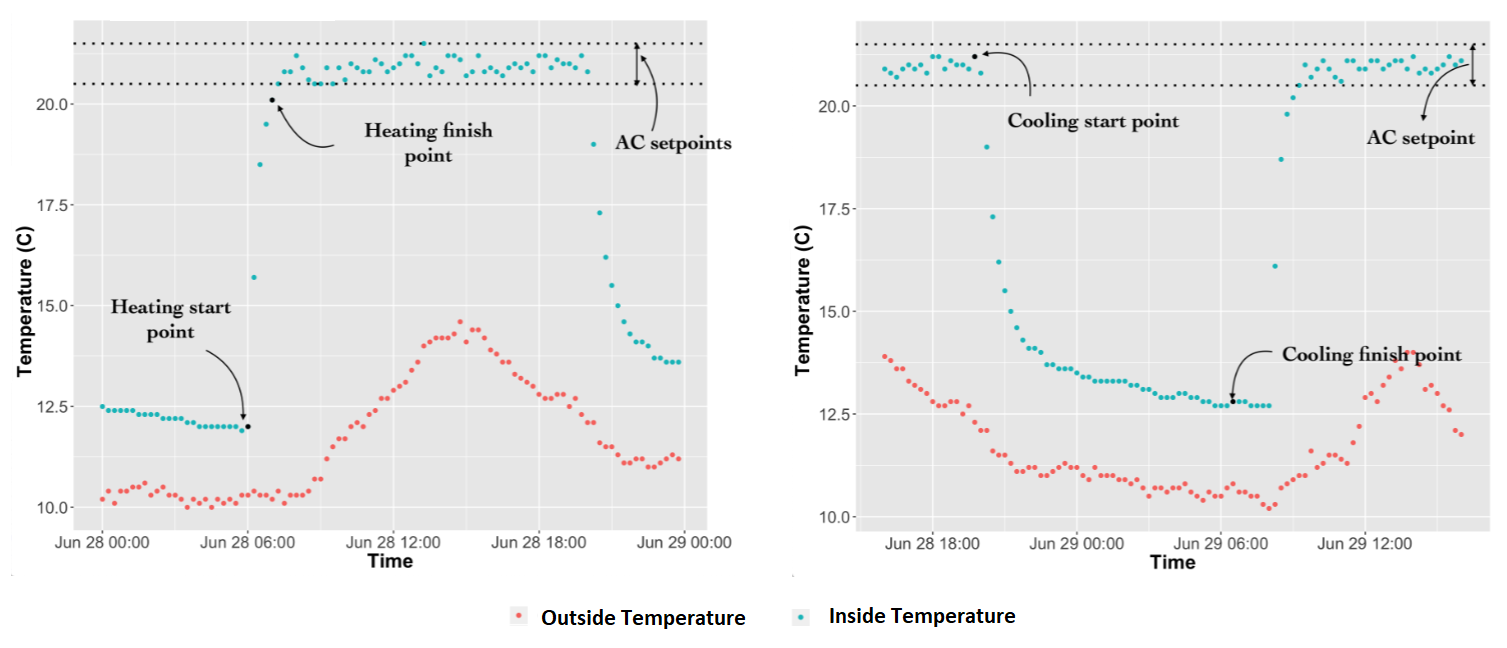}
  \caption{Detection of Heating (Left) and Cooling (Right) Periods }~\label{fig:extract_series}
\end{figure}

Our dataset contains an indicator variable that indicates when the AC system is switched on and off. However, in an exploratory data analysis, we determined that this indicator variable is relatively unreliable in the historic data. Therefore, we opted for a (semi-)manual preprocessing step to extract the active heating and passive cooling periods/series from the original temperature series.
Figure~\ref{fig:extract_series} shows an example for this (semi-)manual detection of active heating and passive cooling periods in between June 28 and June 29 in our dataset. The black points in Figure~\ref{fig:extract_series} show the starting and finishing points of the cooling and heating periods. For the heating section, we only extract the heating period when the temperature almost arrives at the lower setpoint which is 20C in the example. For the cooling section, we pick the whole period when the AC system is completely switched off until it is switched on in the next day.





We extract 68 cooling series and  112 heating series from the lecture theatre temperature data as described above. The corresponding outside temperature series are also extracted as both inside and outside temperatures are used to train the RNN models. 
The extracted heating and cooling series range from 8 to 22 and in order to bring all the temperatures to the same scale, we normalize the data by dividing them using the average temperature during the day time, which is 20 in our case. The purpose of data normalisation is to avoid the placement of RNN outputs within the saturated range \citep{ref_45}. 

Due to the iterative procedure and the simulation through the whole unoccupied period, the models require the corresponding future outside temperatures when predicting the future inside temperatures. These future outside temperatures can be retrieved using a weather forecasting platform. For simplicity, in our experiments we approximate these weather forecasts by using the actual outside temperatures in the dataset. We note that this information feeds into all algorithms compared with each other in the same way, so that it doesn't alter their relative performance or the conclusions of our work.

We build separate RNN models to model heating and cooling using the preprocessed heating and cooling series. Table \ref{tab:series} represents the number of series used to train each model along with the maximum and minimum lengths of the extracted heating and cooling series. 


  

\begin{table}[b]
\begin{center}
\begin{tabular}{ l|c|c|c } 
\hline
\textbf{Type} & \textbf{No: of Series} & \textbf{Minimum Length} & \textbf{Maximum Length}\\ 
\hline
Cooling & 68 & 21 & 51 \\ 
Heating & 112 & 6 & 6 \\ 
\hline
\end{tabular}
\caption{Heating and Cooling Series Information}
\label{tab:series}
\end{center}
\end{table}

We use TensorFlow \citep{ref_18}, an open-source deep-learning platform to implement all RNN models.

\subsection{Error Metric}

In time series forecasting, many error metrics are commonly used, such as Symmetric Mean Absolute Percentage Error (SMAPE) and Mean Absolute Scaled Error (MASE) \citep{ref_52}. However, the primary goal of most of these measures is to evaluate forecasts across series that are on different scales. But in our case, all series are on the same scale, namely temperature in degree Celsius, and hence, we refrain from using specialised forecasting error metrics. Instead we use the Root Mean Squared Error (RMSE), as defined in Equation~\ref{eq:rmse}.

\begin{equation}
    RMSE = \sqrt{\frac{\sum_{i=1}^{N} (a_{i}-f_{i})^2}{N}}\label{eq:rmse}
\end{equation}

In Equation~\ref{eq:rmse}, $N$ is the total number of temperature measurements, $a_{i}$ and $f_{i}$ are the $i$th actual and predicted temperatures, respectively.



\subsection{Benchmark Models}
\label{sec:ben_construction}
We use four state-of-the-art temperature prediction models: SVM, RF, MLR and FFNN as the benchmarks against the RNN to predict inside temperatures. Current inside and outside temperatures are used as inputs to these models to predict future inside temperatures in an iterative manner. 


A grid-search approach is used to tune the hyperparameters of the models. We use 10-fold cross-validation with the training dataset to tune the hyperparameters. 
For SVM, we choose a radial kernel function. The parameters $gamma$ and $C$ (cost) are varied from 1 to 5 with step 0.1 by cross-validation. In the heating section, The best value for $C$ is 0.2, and the chosen value for $gamma$ is 0.1. In the cooling section, the best value for $C$ is 2 and $gamma$ is 0.2.
For RF, we choose 500 for $ntree$, 30 for $ntime$ and 3 for $mtry$ in both cooling and heating sections and change the node size from 1 to 20. For heating and cooling, 2 and 10 are chosen as the node sizes, respectively, with quantile regression computation algorithm.
%
%
For FFNN, the same normalisation method as for the RNN model is used. In FFNN, we choose 0.1 for the threshold, 0.001 for the learning rate and the learning algorithm is back-propagation in both heating and cooling sections. We try different structures of layers and units with (0), (1), (2), (3), (4), (1,1), (1,2), (1,3), (2,3), (2,4), (1,1,2) and (1,2,2). The activation functions that we compare are sigmoid, tanh and logistic. For heating, the best structure is 2-layers structure with 2 and 3 units, respectively, and the activation function is tanh. For cooling, the best model has 1 hidden layer with tanh activation function.

The setpoint optimisation is conducted for all benchmark models in the same way as the RNN model as described in Section \ref{sec:optimisation}, in particular the benchmark models also use the corresponding outside temperatures for training.

\subsection{Evaluation of Prediction Accuracy}
\label{sec:eval_accuracy}

We evaluate our RNN model against the benchmark models based on the prediction accuracy. 
For that, we divide both heating and cooling series into training and test sets, such that 80\% of the series are used for training and 20\% for testing. All prediction models are then trained using the series in the training set. The forecasts are obtained from the trained models for each series in the test set by providing the start inside temperature and corresponding outside temperatures of the series. Finally, the forecasts provided by the separate models are compared with the actual temperatures in the test set and calculate the RMSE for the whole test set.




\begin{table}
\begin{center}
\begin{tabular}{p{5cm}|p{5cm}}
\hline
\textbf{Model}  & \textbf{RMSE}\\ \hline
SVM & 0.72\\ 
RF & 0.75\\ 
MLR & 1.08\\ 
FFNN & 1.89\\ 
RNN & \textbf{0.68}\\ \hline
\end{tabular}
\caption{Comparison of RNN Model with SVM, RF, MLR and FFNN based on RMSE}
\label{tab:rmse}
\end{center}
\end{table}

Table \ref{tab:rmse} represents the RMSE values over the test set, based on the predictions for both cooling and heating series with all considered models. 
From Table \ref{tab:rmse}, it is clear that our proposed RNN model is more accurate in forecasting indoor temperatures compared to the benchmark models.

We further perform a non-parametric Friedman rank-sum test to evaluate the statistical significance of our results in RMSE values. We then use a the Hochberg’s post-hoc procedure \citep{ref_54} to characterise these differences, compared to the best performing model. 
Table \ref{tab:statistical_results} shows the results of the statistical testing with the adjusted $p$-values calculated from the Friedman test with Hochberg’s post-hoc procedure considering a significance level $\alpha = 0.05$. The $p$-value for the overall Friedman rank-sum test is  $5.17 \times 10^{-11}$, that is highly significant. It is clear from Table \ref{tab:statistical_results} that the proposed RNN is the best prediction model that achieves significantly better results than the other prediction models compared against.

\begin{table}
\begin{center}
\begin{tabular}{p{5cm}|p{5cm}}
\hline
\textbf{Model}  & \textbf{$p_{Hoch}$}\\ \hline
RNN & -\\ \hline
FFNN & $4.47 \times 10^{-18}$\\ 
MLR &  $1.09 \times 10^{-5}$\\ 
SVM & $0.025$\\ 
RF & $0.039$\\ \hline
\end{tabular}
\caption{Results of statistical testing}
\label{tab:statistical_results}
\end{center}
\end{table}
 

\section{Case Study: Usage of RNN Model to Optimise the Energy Consumption of HVAC Systems}
\label{sec:case_study}

In this section, we analyse the performance of our proposed RNN model in a real-world scenario of energy optimisation in a university lecture theatre. 
We determine by how much the duration of heating periods can be reduced, as a proxy for energy savings, 
compared with a baseline of static thermal setpoints and the best-performing benchmark from Section~\ref{sec:eval_accuracy}, the SVM model.


\subsection{Baseline Model with Static Thermal Setpoints}
\label{sec:current_model}

\begin{figure}[htb]
\centering
  \includegraphics[width=\columnwidth]{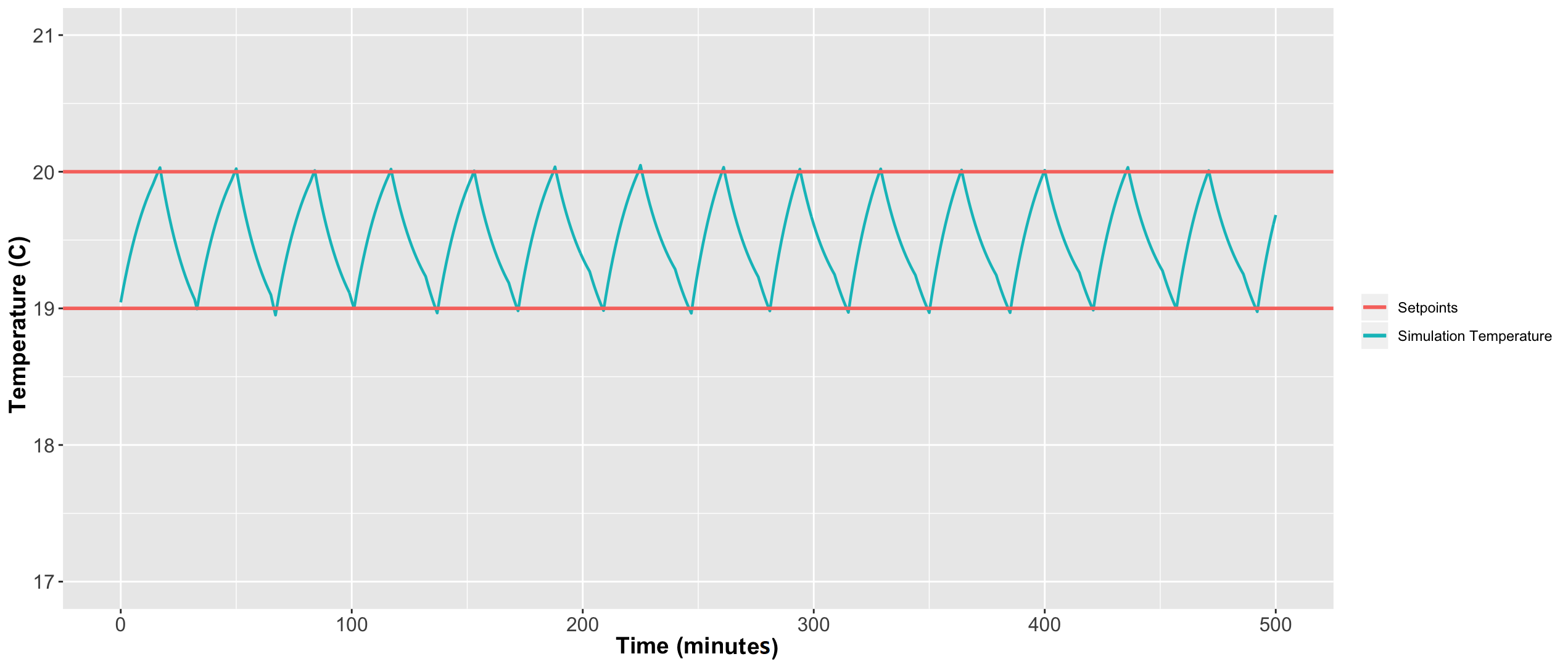}
  \caption{Simulation of Model with Static Thermal Setpoints}~\label{fig:current}
\end{figure}

We assume as a baseline model that the AC system of the lecture theatre uses static thermal setpoints to maintain the inside temperature at a comfortable level. 
Figure \ref{fig:current} shows a simulation of this static thermal setpoint model, with setpoints of 19C and 20C, where the model keeps the inside temperature between the two setpoints irrespective of room occupancy. 

\subsection{Occupation Schedule}
\label{sec:optimised_model}

\begin{figure}[htb]
\centering
  \includegraphics[width=\columnwidth]{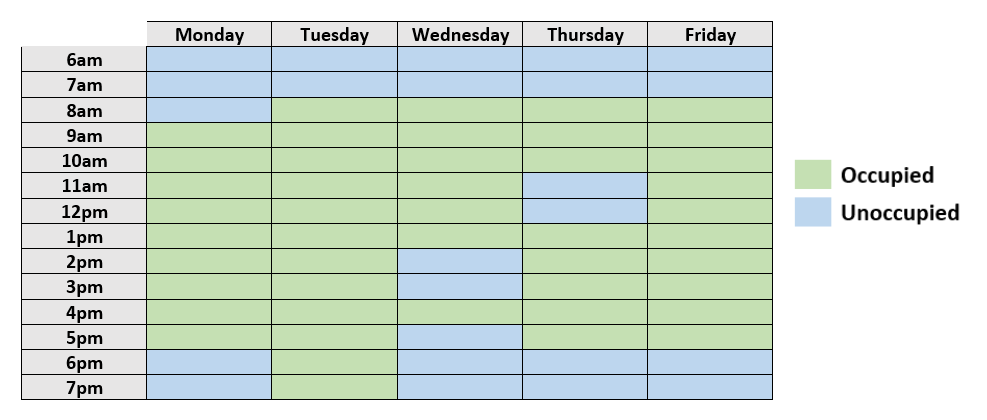}
  
  \caption{Weekly schedule of the lecture theatre in consideration}
  \label{fig:ScheduleRoom}
\end{figure}

Figure \ref{fig:ScheduleRoom} shows the weekly schedule of the lecture theatre in consideration. Though the schedule may change from week to week, the chosen week has a typical occupation pattern and we deem our conclusions drawn from this week representative for a semester. 
We see from the schedule that there are unoccupied periods in the room on Wednesday and Thursday, as well as cases where occupation starts later or finishes earlier than usually. We are able to use this information in our models by reducing the heating time during an unoccupied period, by determining the optimal time points to switch on the AC in the morning, and by switching off the AC in the evening after occupation ends. We optimise our RNN model and the most accurate benchmark, the SVM model as described in Section \ref{sec:optimisation}.

\subsection{Comparison of RNN Model with SVM Model}
\label{sec:comparison}

We compare the models by calculating the reduction percentage of heating time for all days in the schedule. Table \ref{tab:table1} represents the total minutes (total operation time of the HVAC system) required by the models to heat the room with currently used setpoints and optimised models according to the SVM and RNN model simulations along with percentages of the reduction of heating time given by the optimised models. According to Table \ref{tab:table1}, the SVM model and RNN model predict an average of 12.79\% and 20.26\% heating reduction, respectively, over the week. 
As we know from Section~\ref{sec:experiments} that the RNN is significantly more accurate than the benchmarks, and its predicted savings are much higher than the savings from the benchmark model, overall this shows that using the RNN model is superior to using the benchmark.

\begin{table}[b]
\small
\begin{center}
\begin{tabular}{|r|r|r|r|r|r|r|}
\hline
\multicolumn{1}{|c|}{\textbf{Day}} & \multicolumn{3}{|c|}{\textbf{SVM Model}} & \multicolumn{3}{c|}{\textbf{RNN Model}}  \\ \hline
& \textbf{With}   & \textbf{Optimised} & \textbf{Heating} & \textbf{With} & \textbf{Optimised} & \textbf{Heating}  \\
& \textbf{Current}   & \textbf{Model} & \textbf{Reduction} & \textbf{Current} & \textbf{Model} & \textbf{Reduction}\\ 
& \textbf{Setpoints}   &  &  & \textbf{Setpoints} &  & \\ \hline
Monday  &  359 mins &  283 mins & \textbf{21.17\%} &  252 mins &  202 mins & \textbf{19.84\%}  \\ \hline
Tuesday   &  434 mins &  394 mins& \textbf{9.22\%} &  322 mins &  265 mins & \textbf{17.70\%}  \\ \hline
Wednesday   & 563 mins & 464 mins & \textbf{17.58\%} & 396 mins & 284 mins & \textbf{28.28\%}  \\ \hline
Thursday   & 652 mins & 570 mins & \textbf{12.58\%} & 467 mins & 380 mins & \textbf{18.63\%}  \\ \hline
Friday   &  548 mins &  518 mins & \textbf{5.47\%} &  365 mins &  306 mins & \textbf{16.16\%}  \\ \hline
\textbf{Average}  &  \textbf{511.2 mins} &   \textbf{445.8 mins} & \textbf{12.79\%} &  \textbf{360.4 mins} &  \textbf{287.4 mins} & \textbf{20.26\%}  \\ \hline
\end{tabular}
\end{center}
\caption{Comparison of RNN model and benchmark model with the model using the current setpoints of the lecture theatre}
\label{tab:table1}
\end{table}

\begin{figure}[htb]
\centering
  \includegraphics[width=\columnwidth]{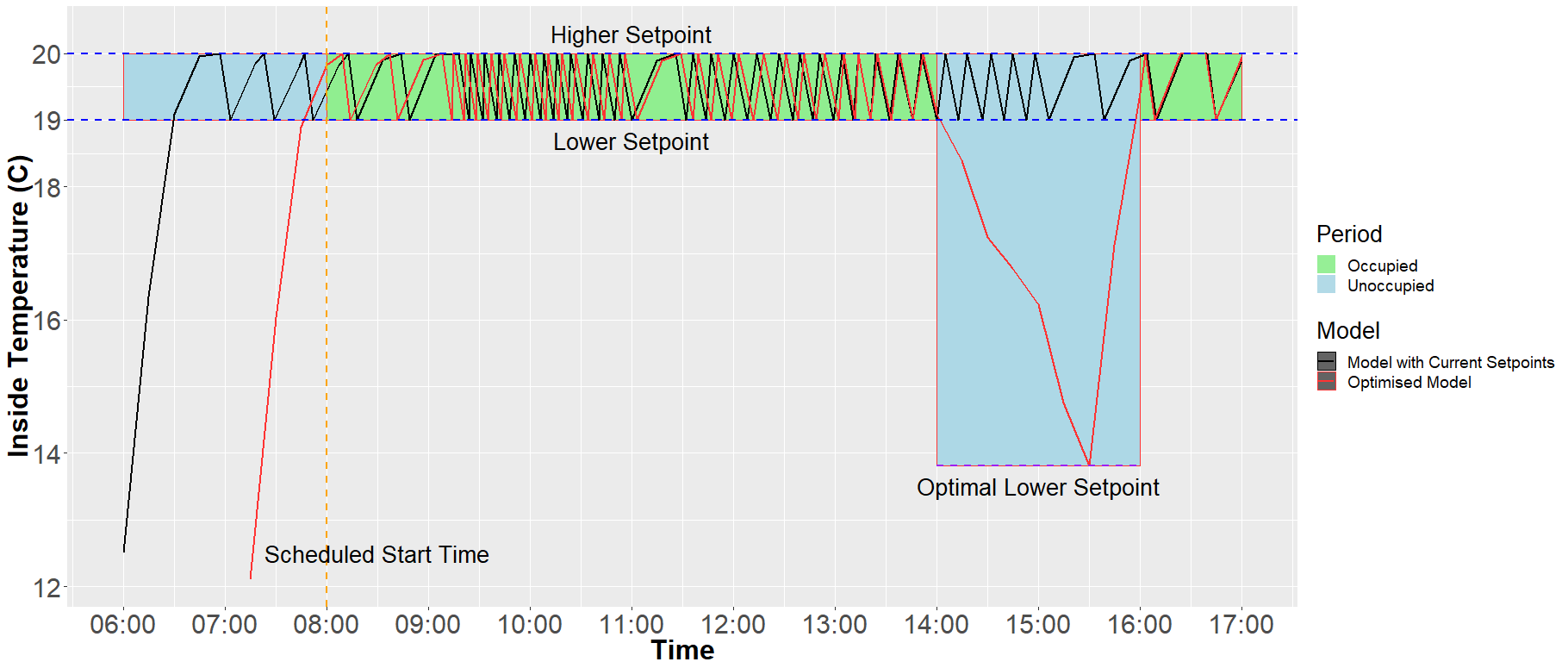}
  
  \caption{Simulation of the RNN Model on Wednesday}~\label{fig:Sep4Direct}
\end{figure}

Figure~\ref{fig:Sep4Direct} shows the simulation of the optimised RNN model and the static baseline, as an example, for Wednesdays. We obtain the optimised curve according to the procedure described in Section~\ref{sec:optimisation}. The behaviour of the model within setpoints depends on its predictions. According to the simulation, the RNN model allows the room to cool down until 13.81C during the unoccupied period and this temperature can be identified as the optimal lower setpoint to be used during that period.

%

For all the days of the week, both the SVM and the RNN model simulations predict a considerable amount of energy saving in terms of heating time reduction compared to the simulation with current setpoints of the lecture theatre in the morning as well as during the unoccupied periods in the afternoon. The RNN model simulation predicts a higher amount of energy saving than the SVM model simulation. The RNN model allows the room to cool down to a lower temperature during the unoccupied periods of the room and as a result, it predicts a higher energy saving compared to the SVM model on Wednesdays and Thursdays, according to our simulations. Furthermore, the simulations of the RNN model allow the room to start heating much closer to the scheduled start time on each day compared to the SVM model simulation and therefore, it predicts an energy saving more than 16\% on Monday, Tuesday and Friday even though the room does not contain any unoccupied periods in the afternoon.

\section{Conclusion}
\label{sec:conclusion}
Many buildings use static thermal setpoints to maintain their inside temperatures at a comfortable level irrespective of building occupancy. This causes energy wastage and increases energy related expenses.
In this paper, we have proposed a deep-learning model based on RNNs to predict the inside temperatures of a particular room. RNNs are particularly useful in our setup as they are able to learn across many relatively short time series, which in our case allows us to train separate models on particular operation modes. We have compared the prediction accuracy of our RNN model with four state-of-the-art temperature prediction models based on SVM, RF, MLR and FFNN. We have been able to show that the RNN is significantly more accurate in predicting indoor temperatures than the benchmark models. 
We have analysed the usage of our proposed RNN temperature prediction model and the best-performing benchmark method in a real-world scenario by using it to optimise the energy consumption of the HVAC system of a lecture theatre.
We evaluate these models against simulations of the current AC system of the lecture theatre. The proposed model predicts considerable potential energy saving in terms of reducing the time period required to heat the lecture theatre, above both the benchmark and the current system.
As the RNN model provides more accurate temperature forecasts as well as predicts more energy saving, it can be identified as an appropriate way for predicting temperatures.

The success of this approach encourages as future work to build a global temperature prediction model which can predict the future temperatures related to any room type. Different rooms have different characteristics such as size, occupancy and number of windows. Therefore, developing a prediction model that can predict the temperatures belonging to any room type will be useful. Developing a temperature prediction model using ensemble mechanisms is another possible approach that would further increase the performance of our optimised RNN model.
Furthermore, our research is also applicable to similar room settings outside the university space, e.g., hotel conference rooms, libraries, and theatres, which highlights to potential commercial benefits of our research.

\section*{Acknowledgement}
We thank Honeywell International Inc for supporting this research, and Buildings and Property Division, Monash University, Australia for providing the temperature data to conduct the experimental study.






 \bibliographystyle{elsarticle-harv}

\bibliography{sample}

\end{document}